\documentclass[lettersize,journal]{IEEEtran}

\usepackage{amsmath,amsfonts}
\usepackage{algorithmic}
\usepackage{algorithm}
\usepackage{array}
\usepackage[caption=false,font=normalsize,labelfont=sf,textfont=sf]{subfig}
\usepackage{textcomp}
\usepackage{stfloats}
\usepackage{url}
\usepackage{verbatim}
\usepackage{graphicx}
\usepackage{cite}
\usepackage{color}
\usepackage{flushend}
\usepackage[utf8]{inputenc}
\usepackage[switch]{lineno}

\begin{document}
\newcommand{\jiaxin}[1]{\textcolor{red}{#1}}
\title{Enhancing Lane Segment Perception and Topology Reasoning with Crowdsourcing Trajectory Priors}

\author{Peijin Jia$^{1}$$^*$, Ziang Luo$^{1}$$^*$, Tuopu Wen$^{1}$,  Mengmeng Yang$^{1}$, Kun Jiang$^{1}$, Le Cui$^{2}$, Diange Yang$^{1}$
\thanks{$^{1}$Peijin Jia, Ziang Luo, Kun Jiang, Mengmeng Yang, Tuopu Wen, Diange Yang are with the School of Vehicle and Mobility, Tsinghua University, Beijing, 100084, China.}
\thanks{$^{2}$Le Cui is with DiDi Chuxing.}
\thanks{$^{*}$Equal contribution.}
\thanks{This work was supported in part by Beijing Municipal Science and Technology Commission (Z241100003524013, Z241100003524009), the National Natural Science Foundation of China(U22A20104, 52472449, 52402499), and Beijing Natural Science Foundation(L231008, L243008), the State-funded postdoctoral researcher program of China (GZC20231285)
, China Postdoctoral Science Foundation (2024M761636). This work was also sponsored by Tsinghua University-Didi Joint Research Center.}
\thanks{Corresponding author: Diange Yang, Kun Jiang, Mengmeng Yang and Tuopu Wen}
}




\maketitle

\begin{abstract}
In autonomous driving, recent advances in online mapping provide autonomous vehicles with a comprehensive understanding of driving scenarios. Moreover, incorporating prior information input into such perception model represents an effective approach to ensure the robustness and accuracy. However, utilizing diverse sources of prior information still faces three key challenges: the acquisition of high-quality prior information, alignment between prior and online perception, efficient integration. To address these issues, we investigate prior augmentation from a novel perspective of trajectory priors. In this paper, we initially extract  crowdsourcing trajectory data from Argoverse2 motion forecasting dataset and encode trajectory data into rasterized heatmap and vectorized instance tokens, then we incorporate such prior information into the online mapping model through different ways. Besides, with the purpose of mitigating the misalignment between prior and online perception, we design a confidence-based fusion module that takes alignment into account during the fusion process. We conduct extensive experiments on OpenLane-V2 dataset. The results indicate that our method's performance significantly outperforms the current state-of-the-art methods. Code is released is at \url{https://github.com/wowlza/TrajTopo}
\end{abstract}


\section{Introduction}
\IEEEPARstart 
In autonomous driving systems, high-definition (HD) map plays a critical role by providing accurate geographical map elements and comprehensive topology information. Nevertheless, the extensive adoption of HD map is frequently impeded by the high costs of annotating the maps and the difficulty of updating. Hence, there is a growing interest in the potential of online local map construction with onboard sensors to supplant the conventional process of creating offline HD map.

With the advancement of online mapping technology, researchers have observed that online map is susceptible to environmental influences due to the inherent limitations of onboard sensors. To tackle this issue, recent research\cite{luo2024augmenting}, \cite{xiong2023neural}, \cite{gao2024complementing} have sought to leverage prior information such as Standard Definition (SD) map, local history map, satellite map as effective supplement.

\begin{figure}[ht!]
    \centering
    \includegraphics[width=0.5\textwidth]{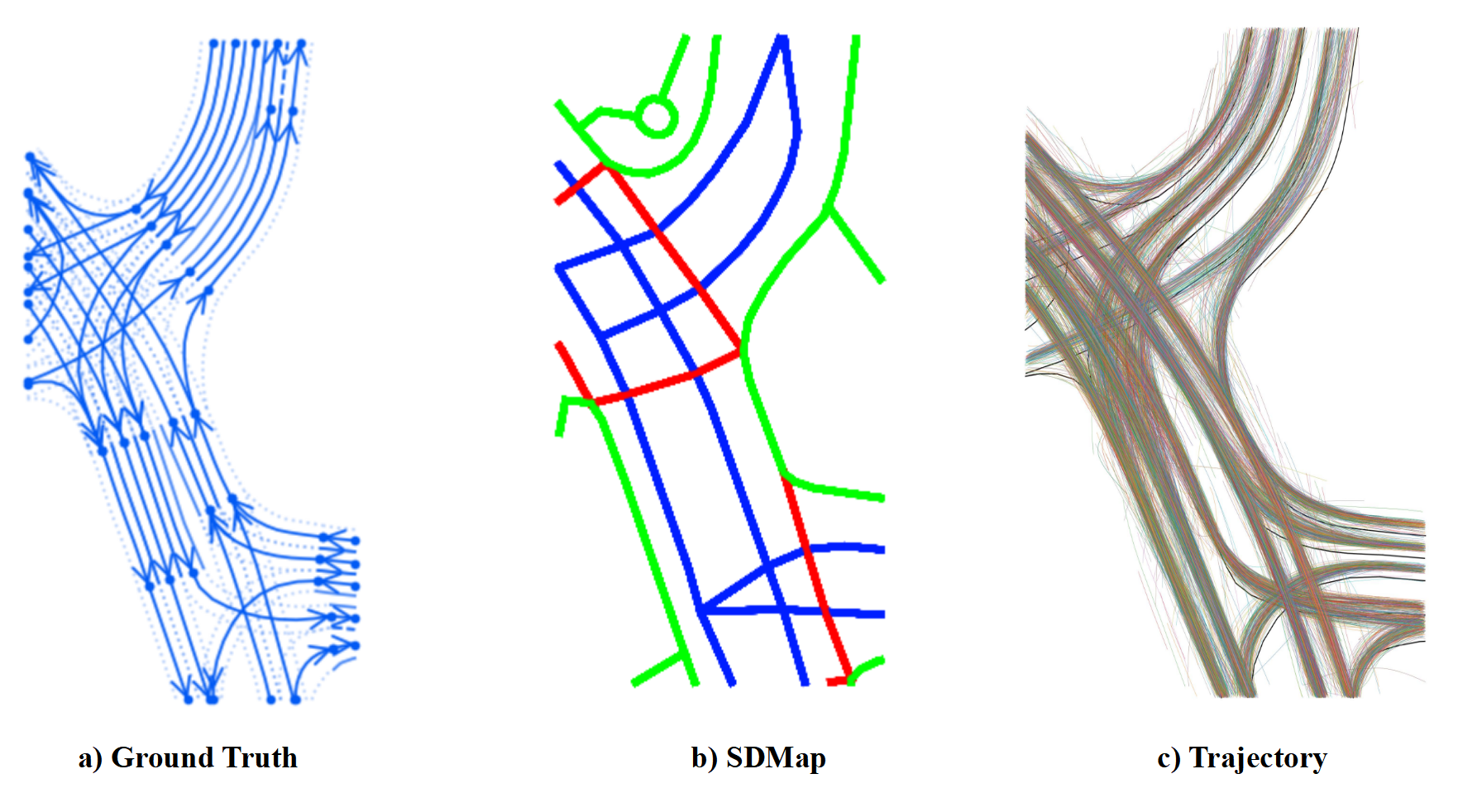}
    \caption{Comparison of Different Priors: a) Ground Truth. b) SDMap, where blue lines represent roads, green lines represent sidewalks, and red lines represent crosswalks. c) Trajectory. The trajectory data is more closely match the geometric structure of lanes, especially in the intersection area.}
    \label{fig:comparison_prior}
\end{figure}

However, there are still challenges associated with the utilization of prior information, such as which category of prior information is more accessible and better reflects the road structure, how to integrate the prior information into model and solve the misalignment problem in space and semantics.

As depicted in Figure \ref{fig:comparison_prior} b), the most commonly used prior information SD map outlines the framework of the road structure, yet it only reflects the road-level map. This observation prompts us to consider whether there exists prior information that aligns more closely with the lane-level road structure. As crowdsourcing trajectory data direct reflect the natural driving behavior following lane connections and centerline paths, it can provide more detailed information for map prior. As shown in Figure \ref{fig:comparison_prior} c), the trajectory data closely match the geometric structure of lanes. 
Moreover, crowdsourcing trajectory data is broadly covered, easily accessible, and frequently updated. 
Given the aforementioned considerations, we augment lane segment perception with trajectory data as supplementary prior information to compensate for sensor data insufficiency.


Specifically, we initially extract trajectory data from Argoverse2 motion forcasting Dataset\cite{wilson2023argoverse}. To effectively integrate trajectory priors, we preprocess the data, converting it into formats suitable for neural network training. We encode trajectory using two primary methods: 1) Rasterzied Heatmap, which involves transforming data into a heatmap by evaluating grid density and direction. 2) Vectorized Instance Tokens, which extract the most representative trajectory by sampling or clustering. 

The subsequent challenge is to fully exploit this trajectory data. We explore the effects of applying prior information to enhance the Bird's Eye View (BEV) feature and refine the queries and reference points of lanesegment decoder.

To address misalignment of space and semantics between prior information and online perception, we further design an alignment module that integrates spatial position correction and semantic confidence fusion, using BEV segmentation as a supervision.

We conduct extensive experiments to compare different fusion methods. Results demonstrate that trajectory data significantly enhances model performance in online mapping tasks, regardless of the fusion scheme used. The best fusion method surpasses state-of-the-art methods by a notable margin, achieving gains of $+10.0$ and $+9.25$ on mAP and topology metrics, respectively. 


In summary, our contributions are as follows:
\begin{itemize}
\item We extract numerous trajectories from motion forecasting dataset and align them with OpenLane-V2 for regional mapping, using them as additional prior information. We plan to open-source this supplementary trajectory dataset later.
\item We examine various fusion strategies and design a novel fusion framework. We identify the optimal fusion method and develop an alignment module to mitigate misalignment in prior information.
\item 
Our model surpasses the current state-of-the-art methods, proving the effectiveness of the trajectory priors and the designed modules.
\end{itemize}

\begin{figure*}[ht!]
    \centering
    \includegraphics[width=0.8\textwidth]{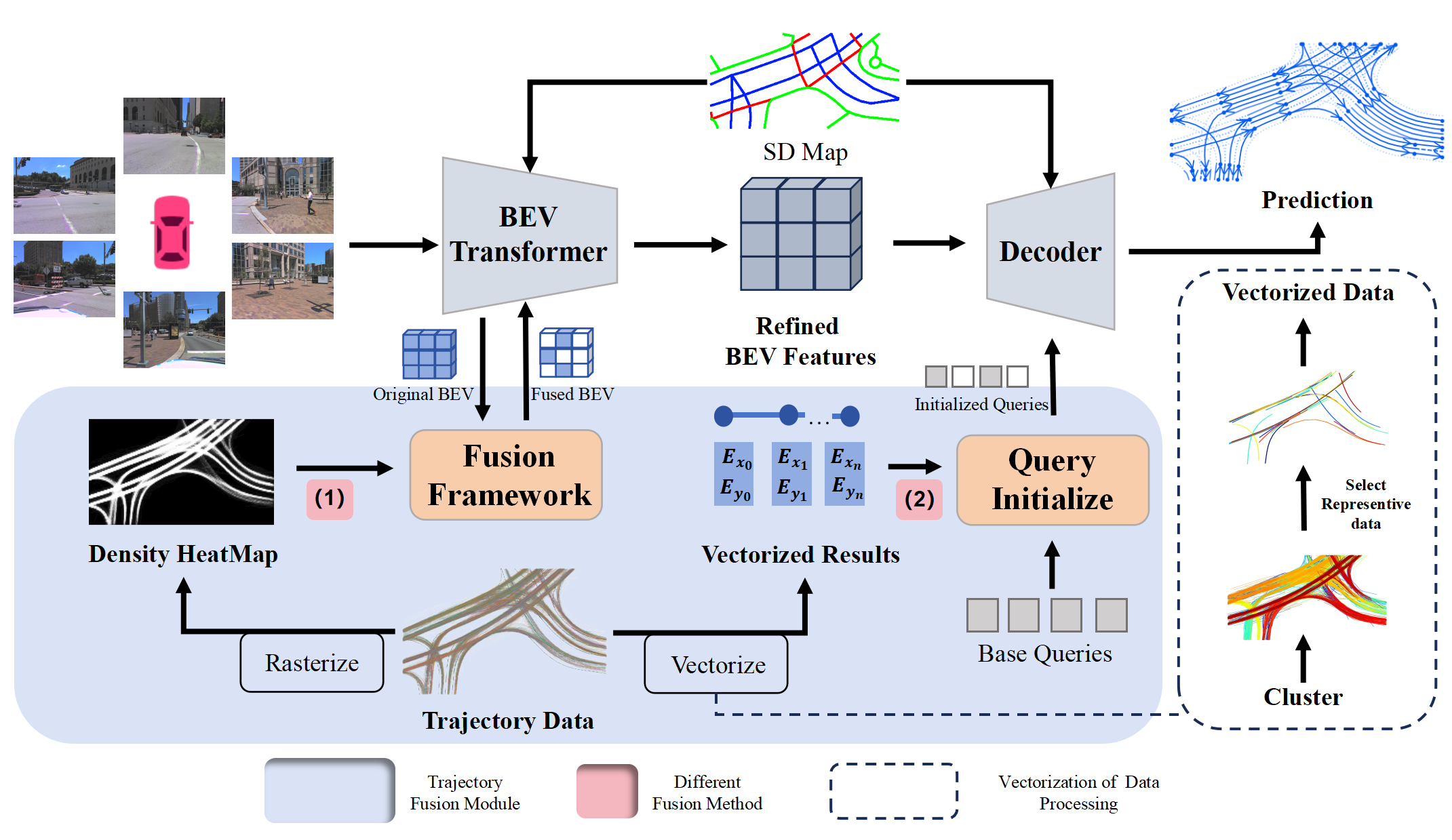}
    \caption{\textbf{The overview model architecture.} The model receives surrounding images, a local aligned SDMap and trajectory data as prior inputs and the model follows the typical encoder-decoder paradigm. We add the sdmap upon previous work. Then we innovatively represent trajectory data in two forms and compare different fusion strategies, ultimately identifying the optimal fusion strategy.}
    \label{fig:overview}
\end{figure*}

\section{Related Work}
\subsection{Topology Reasoning}
As online mapping is becoming increasingly popular, an increasing number of researchers have begun to generate lane-level topological relationships directly based on onboard sensor information. STSU\cite{can2021structured} is a pioneering work in online road topology reasoning, it takes a front view picture as input and infers topological relationships based on the detected centerline to obtain the final lane map. Building on this foundation, TPLR\cite{can2022topology} represents road topology with a set of directed lane curves, introducing the concepts of minimal cycles and their covers. RoadPainter\cite{ma2024roadpainter} extracts a set of points from each centerline mask to improve the accuracy of centerline prediction. In addition to modeling the lane map with centerlines as nodes, LaneGAP\cite{liao2023lane} proposes to reason topological relationships with entire lanes as nodes, while LaneSegNet\cite{li2023lanesegnet} introduces lane segment perception. It detects road topological structures with lane segments that include both lane centerlines and lane boundaries as nodes. 

Beyond topological reasoning through graph network construction, Tesla proposes the use of language model for topological structure inference on AI Day, 2022\cite{tesla2022}. RoadNet\cite{lu2023translating} and LaneGraph2Seq\cite{peng2024lanegraph2seq} also employ language model to model topological structures, encoding details related to lane lines, such as node positions, attributes and geometric parameters of lane line spline curves. They then utilize an autoregressive decoder for inference, employing sequential processing techniques to generate the geometric topology of lane lines.

Building on this foundation, OpenLane-V2\cite{wang2024openlane} dataset incorporates traffic signals into the topological relationships, and introduces the TOP evaluation metric for road topology structure detection, launching a challenge for this task. TopoNet\cite{li2023graph} is the first end-to-end network to incorporate traffic signals into topological reasoning, the model first uses an embedding network to extract semantic information of traffic elements, then transforms them into a unified feature space for matching with centerlines, and subsequently uses a graph neural network to infer topological relationships.
TopoMask\cite{kalfaoglu2024topomaskv2} designs instance-level queries during the centerline detection process to generate masks, which are then used to create point sets based on instances. TopoMLP\cite{wu2023topomlp}, on the other hand, is designed for the prediction head of topological relationships. It embeds the predicted lane point coordinates through an MLP and then integrates them into the lane query feature to fuse distinctive lane information. 

\subsection{Map Prior}
Prior information can effectively enhance the robustness of models and reduce the uncertainty brought by vehicle onboard sensors. Many studies have attempted to introduce map priors to generate online high-precision map data, which can be broadly categorized into explicit forms such as standard map information and implicit forms like temporal information. In terms of implicit priors, NMP\cite{xiong2023neural} provides autonomous vehicles with long-term memory capabilities by fusing past map prior data with current perception data. MapPrior\cite{zhu2023mapprior} combines a discriminative model with a generative model, generating preliminary prediction results based on the existing model during the prediction phase, and encoding these as priors into the discrete latent space of the generative model. StreamMapNet\cite{yuan2024streammapnet} utilizes temporal information by propagating queries and fusing BEV features. While DiffMap\cite{jia2024diffmap} uses diffusion model to refine the BEV feature, capturing the map structure prior.

In the utilization of explicit priors, SatForHDMap\cite{gao2024complementing} uses satellite images as prior information to build high-precision maps in real-time. MapLite 2.0\cite{ort2022maplite} pioneers the introduction of Standard Definition (SD) map with a SLAM approach. It formulates the problem as a maximum likelihood estimation and then uses sensor measurements to update the decision variable set of the HDMap. P-MapNet\cite{wang2024priormapnet}, MapEX\cite{sun2023mind}, and SMERF\cite{luo2024augmenting} introduced SD Map into online mapping algorithms under the DETR framework. P-MapNet\cite{jiang2024p} stores SD map in a rasterized format, while SMERF\cite{luo2024augmenting} designs a Transformer-based SD Map encoder and fuses the prior information with the BEV feature. MapEx\cite{sun2023mind} concatenates point coordinates from the vectorized SD Map with the corresponding one-hot vectors of road type, encodes them directly, and initializes the queries in the decoder with these features.
The utilization of SD Map has been well-explored, but the inherent bias in SD Map also limits the ceiling of the enhancement effect. In this paper, we explore prior information that is more aligned with the lane segment perception task--trajectory, to further improve the model's detection performance. Additionally, Autograph\cite{zurn2023autograph} attempts to incorporate trajectories (from Argoverse 2) for topology reasoning; however, it is an offline method that relies on aerial imagery for inference.

\section{Supplementary Trajectory Dataset}
The OpenLane-V2\cite{wang2024openlane} dataset represents a pioneering effort in the realm of topology reasoning for traffic scene structure. It is built upon the Argoverse2\cite{wilson2023argoverse} dataset, with downsampling and additional annotations applied. 



 To augment the OpenLane-V2 with trajectory prior, we investigate the trajectory data present in the Argoverse2 Motion Forecasting Dataset. Although the motion forecasting dataset and the dataset used for perception tasks differ in scenarios and time, they are collected from the same urban areas.
Therefore, for the $6$ cities under consideration, we determine the pose of the map elements by leveraging the vehicle's pose information and merge this with the extracted trajectory data. By employing this method, we have identified the historical trajectory prior data corresponding to each scenario.

Besides, we find that the proportion of frames containing trajectory data that is five times the number of lane segments is remarkably high, accounting for $88.63\%$ of the training dataset and $86.18\%$ of the validation dataset, revealing that the quantity of trajectory data is sufficient to meet the requirements for trajectory priors. 


To rigorously assess data quality, we implement centerline-aligned Intersection over Union (IoU) for comprehensive quantitative evaluation. The centerline annotations are projected onto BEV maps with a line width of 0.75 m, and the intersection ratio is then computed. Our results show that SD Map achieves an IoU of 0.16, whereas the trajectory data attains an IoU of 0.39, highlighting the superior geometric fidelity of the trajectory data.



\section{Methodology}
Having acquired the complementary dataset of trajectory data, this section proceeds to detail the data processing for trajectories and introduces the fusion framework.
\subsection{Overall Architecture}
The comprehensive architecture is illustrated in Figure \ref{fig:overview}. Building on the foundation of the lane segment perception task\cite{li2023lanesegnet}, our model processes surrounding images, a locally aligned SD Map, and novel trajectory data. It then detects the geometries of lane segments and generates their topological relationships. 

As the topology detection module proposed by LaneDag\cite{jia2024lanedag} performs better, we adopt it as the topology head.
In the SD Map fusion module, we enhance SD Map integration like SMERF \cite{luo2024augmenting} and add our alignment module. 

For the trajectory fusion module, we introduce an innovative representation of trajectory data in two formats: rasterzied heatmap and vectorized instance tokens. After designing and evaluating different fusion strategies, we identify the optimal approach to integrate trajectory data. Details are illustrated in the following part.

\subsection{Data Processing for Trajectories}
The raw trajectory data varies in quality and lacks self-verifiable reliability, making it unsuitable as prior information. To improve usability, we first filter the data by trajectory length and apply an averaging filter to reduce random fluctuations.

Despite filtering and smoothing, each frame still contains thousands of trajectories, making it challenging for the neural network to process. To address this, we apply two distinct encoding methods to enable the network to effectively learn and utilize the trajectory priors.
\subsubsection{\textbf{Rasterzied Heatmap}}
To rasterize trajectory data, we analyze the midpoints and angles between consecutive trajectory points, mapping them onto a grid-based coordinate system. For each grid cell intersected by a trajectory, we record both the visit frequency and movement direction, ensuring a structured representation of the trajectory data.

Mathematically speaking, let the set of trajectory be defined as: $
\mathbf{T}=\{ \mathbf{P}^{(1)}, \ldots, \mathbf{P}^{(m)} \}$
where each trajectory \(\mathbf{P}^{(i)}\) is a sequence of \(n\) two-dimensional points $\mathbf{P}^{(i)} = \{\mathbf{p}_1^{(i)}, \mathbf{p}_2^{(i)}, \ldots, \mathbf{p}_n^{(i)}\}$. For each pair \( \mathbf{p}_i\) and \( \mathbf{p}_{i+1}\), the direction angle 
\( \theta_{i,i+1} = \arctan(y_{i+1} - y_i, x_{i+1} - x_i) \).

Create an \( H \times W \) grid to segment the space. Each grid cell represents $\Delta x, \Delta y$ in real world. Then we calculate the number \( N \) and the average direction angle \( \theta \) for the set of trajectory segments passing through each grid. \( N_{max} \) is the max passing trajectory number among all grids.




After getting density and direction, we normalize the density values to the interval $(0,1)$. The direction values transform to the interval \( \left(-\frac{\pi}{2}, \frac{\pi}{2}\right) \) using an arctangent function.
 
This process produces a heatmap consisting density and direction information.

\subsubsection{\textbf{Vectorized Instance Tokens}}
Another approach involves vectorizing the trajectory data, which aligns well with current data processing trends. Given the large volume of trajectory data, our goal is to extract the most representative samples to support neural networks in learning their underlying patterns effectively. To achieve this, we employ two strategies: the K-means clustering algorithm and the Farthest Point Sampling (FPS) method.

\textbf{Clustering Algorithm}
Numerous studies have explored diverse clustering algorithms for the analysis of trajectory data, including TRACLUS\cite{lee2007trajectory} and RoadUserPathways\cite{kaths2024crossing}. 
However, these methods all exhibit significant time complexity. Given the vast volume of trajectory data we are dealing with, we ultimately opt for the K-Means\cite{krishna1999genetic} clustering algorithm due to its efficiency in handling large datasets.

K-means clustering algorithm is a classic center-based clustering technique aimed at partitioning trajectory data into \( K \) clusters by minimizing the sum of the distances from data points to their cluster centers. Given a set of trajectory samples $T$,
translate it into a two-dimensional matrix \( \mathbf{A}^* \), where \( \mathbf{A}^* \in \mathbb{R}^{m \times n} \), with \( m \) representing the number of trajectories and \( n \) representing the number of feature per trajectory (the dimension of the coordinate points).


We initialize by selecting \( K \) trajectories as the initial cluster centers \( \{\mathbf{\mu}^{(1)}, \ldots, \mathbf{\mu}^{(k) }\}\). For each trajectory \( \mathbf{P}^{(i)} \), we calculate its distance to each cluster center and assign it to the nearest cluster center:
\begin{equation}
c^{(i)} = \arg\min_{j} \, d(\mathbf{P}^{(i)}, \mu^{(j)}), j \in 1,2,...,k
\end{equation}
where \(d(\mathbf{P}^{(i)}, \mu^{(j)})\) is the distance function from trajectory \(\mathbf{P}^{(i)}\) to trajectory $\mu^{(j)}$ and \( c^{(i)} \) is the cluster index to which the trajectory \( \mathbf{P}^{(i)} \) is assigned. The center of each cluster \( k \) is then updated to the mean of all trajectories assigned to that cluster.

The algorithm iterates until either the maximum number of iterations is reached or the change in cluster centers falls below a tolerance of 0.0001. Finally, we designate the cluster centers as representative trajectories, forming the final set of trajectory clusters.



\textbf{Farthest Point Sampling}
Thousands of trajectories, akin to lidar point clouds, are characterized by their sparsity and vast quantity. In traditional point cloud processing methods, researchers often employ Farthest Point Sampling (FPS) to condense feature effectively. The algorithm is a greedy procedure for selecting a subset of points such that the points in \(\mathcal{S}\) are as dispersed as possible. Consequently, we have adopted the Farthest Point Sampling (FPS) method, augmented by the Frechet distance, as an additional strategy to enhance our analysis.

We iteratively select a trajectory that maximizes the Frechet distance to the current subset \(\mathcal{S}\) and then add the selected trajectory into subset. This is formalized as:
\begin{equation}
s = \mathop{\arg\max}\limits_{i} \, d\left( (\mathbf{T} - \mathcal{S})^{(i)}, \mathcal{S} \right), \quad i \in \left[ 1,\, N_p - |\mathcal{S}| \right]
\end{equation}

where $d\left((\mathbf{T} - \mathcal{S})^{(i)}, \mathcal{S} \right)$ is the distance function from trajectory $(\mathbf{T} - \mathcal{S})^{(i)}$ to any trajectory in the subset \(\mathcal{S}\). Here, $N_p$ is the number of the complete trajectory set $\mathbf{T}$, $|\mathcal{S}|$ indicates the size of the currently sampled subset, and \( s \) is the selected index of trajectory in $\mathbf{T} - \mathcal{S}$.

This process continues until \(\mathcal{S}\) contains a certain amount of trajectories, ensuring a uniform sampling that captures the global structure of the dataset. 

\subsection{Prior Fusion Framework}

Given the high-quality trajectory information obtained, determining how and where to incorporate this prior information is crucial for optimal model performance. The introduction of prior information implies greater uncertainty because of the misalignment of prior information. Thus we propose incorporating the prior during BEV feature construction to reduce negative effects from sensor-map inconsistencies and design an alignment module. 

\subsubsection{\textbf{Fusion Strategy}}


We employ a BEVFormer-based encoder \cite{li2022bevformer} to generate BEV features and a DETR-based map decoder to detect up to $N$ map elements. The encoder converts multi-camera features into a BEV representation, while learned BEV features are input into a transformer map decoder. The decoder employs cross-attention to combine query embedding with BEV features. Decoded queries are then transformed into map coordinates through linear layers.

When using prior information to enhance BEV feature, there are three widely-used methods to integrate prior information: addition, concatenation, and cross-attention mechanisms. Considering the heatmap and bev features both are the rasterzied format, we adopt addition mechanism. During BEV feature learning stage, we merge the rasterized trajectory prior through element-wise addition (as shown in Figure \ref{fig:overview} method(1)).

And the prior information can serve the initialization of decoder queries and reference points, where we generate EX queries to encapsulate vectorized trajectory information and using the coordinates of the trajectory as the initial reference points of EX queries (Figure \ref{fig:overview} method(2)). These EX queries are then combined with conventional learnable queries to meet the required number of queries. This comprehensive set of queries is subsequently fed into a transformer decoder, where they are converted into predictions via linear layers.

By combining these three fusion methods across two stages, we can devise various fusion strategies. The diagrams in Figure \ref{fig:overview} illustrate these fusion strategies, and the most effective one, determined through extensive experimentation, is discussed in detail in Section \ref{sec:expriment}.


\begin{figure}[t!]
    \centering
    \includegraphics[width=0.5\textwidth]{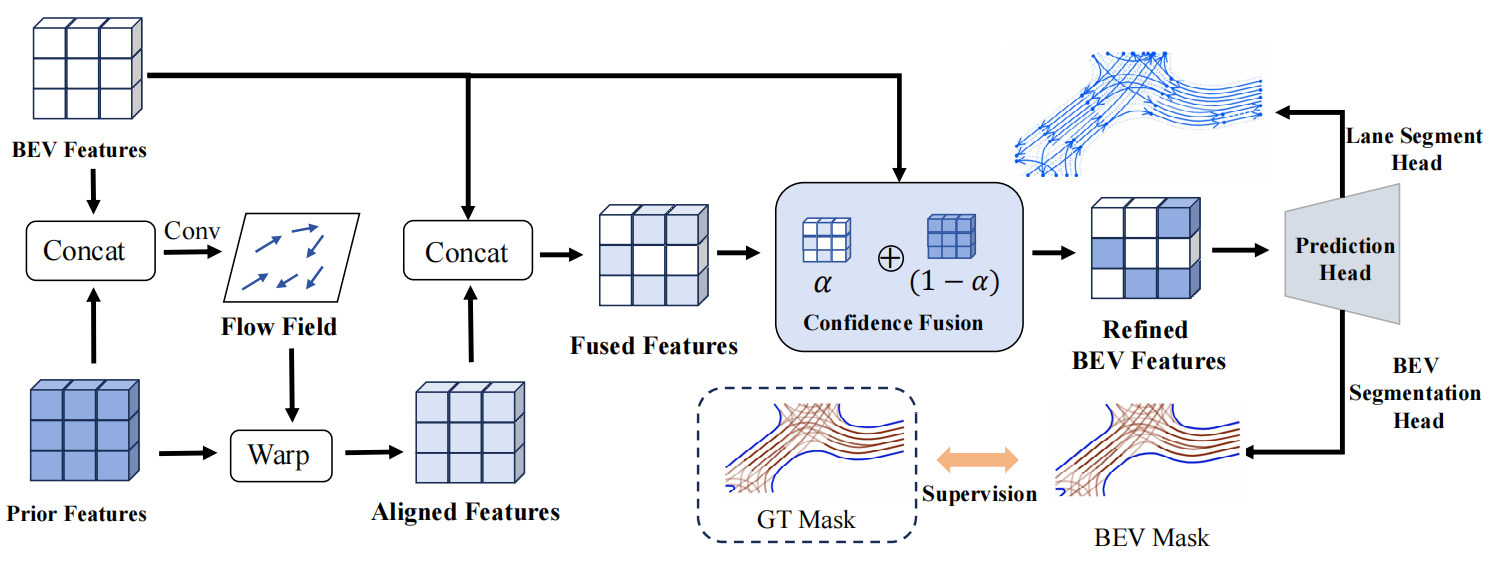}
    \caption{\textbf{Illustration of alignment Module:} We initially concatenate the prior feature with BEV feature and feed them into several convolutional layers to predict the coordinate offsets at each position. Subsequently, we employ a warp operation to achieve spatially aligned feature. These aligned feature then proceed through the confidence fusion module to further integrate the prior information. Throughout the training process, BEV segmentation supervision is implemented to correct misalignments in both space ans semantics.}
    \label{fig:integrity_monitor}
\end{figure}

\subsubsection{\textbf{Alignment Module}}
To mitigate spatial bias and uncertainty in rasterized data, we developed a specialized fusion alignment module.

This module includes three main components: the Confidence Fusion Module, the Spatial Alignment Module, and BEV (Bird’s Eye View) segmentation supervision. The Confidence Fusion Module assigns adaptive weights to prior data, inspired by \cite{liu2019learning}, enhancing the model’s decision-making in uncertain and noisy conditions. The Spatial Alignment Module ensures precise integration and alignment of prior data. BEV segmentation supervision provides rich contextual information, helping to learn the spatial bias and confidence parameters. These components collectively enhance vehicle perception, enabling more accurate and reliable operation. Figure \ref{fig:integrity_monitor} illustrates the architecture of the alignment module.

To address discrepancies between trajectory data and the generated BEV feature caused by localization errors, we ues a Spatial Alignment Module. We first concatenate the prior feature \(\mathcal{F}_{prior}\) with the BEV feature \(\mathcal{F}_{BEV}\) and pass them through multiple convolutional layers to predict the coordinate offsets \(\Delta_{hw} \in \mathbb{R}^{H \times W \times 2}\) at each spatial position, askin to the flow model\cite{li2020semantic}. Each position \((h, w)\) is then mapped to a new location using a warp operation:

\begin{equation}
\mathbf{F}_{prior}'(h, w) = \sum_{(h', w') \in \mathcal{N}(h, w)} w_{p} \cdot \mathcal{F}_{prior}(h', w')
\end{equation}

where \((h', w')\) represents the neighboring pixels (top-left, top-right, bottom-left, bottom-right) around the position \((h + \Delta_{hw1}, w + \Delta_{hw2})\), and \(w_p\) are the bilinear interpolation weights. Here, \(\Delta_{hw1}\) and \(\Delta_{hw2}\) denote the learned 2D offsets for position \((h, w)\). This interpolation approach ensures that the aligned feature \(\mathbf{F_{prior}}'(h, w)\) is accurately adjusted for localization errors.


Moreover, after encoder processing, we get the BEV feature \(\mathcal{F}_{BEV}\)  and prior feature \( \mathcal{F}_{prior}'\). The prior feature is then upsampled to match the shape of the BEV feature.

We introduce two spatial importance weight matrices, \( \alpha^l \) and \( \beta^l \), which are adaptively learned by the network. The feature map \( y^l \) is generated using:
\begin{equation}
\mathbf{y}^l = \alpha^l \odot \mathcal{F}_{BEV} + \beta^l \odot \mathcal{F}_{prior}'
\end{equation}
where the constraints \( \alpha^l + \beta^l = 1 \) and \( \alpha^l, \beta^l \in [0, 1] \) ensure normalized weights.

To derive these weights, a $1 \times 1$ convolution kernel is applied to the original feature map, producing \( \lambda^l_\alpha \) and \( \lambda^l_\beta \). The final weights are computed using the softmax function::
\begin{equation}
\alpha^l = \frac{e^{\lambda^l_\alpha}}{e^{\lambda^l_\alpha} + e^{\lambda^l_\beta}},
\beta^l = \frac{e^{\lambda^l_\beta}}{e^{\lambda^l_\alpha} + e^{\lambda^l_\beta}}
\end{equation}
The softmax function ensures \( \alpha^l \) and \( \beta^l \) lie between 0 and 1, with their sum equal to 1, achieving normalized weighting. Through the Confidence Fusion Module, the model dynamically merges BEV feature \( \mathcal{F}_{BEV} \) with prior information \( \mathcal{F}_{prior}' \) based
on the network-learned weights, enhancing the model's evaluation of prior information and its adaptability across various scenarios.

Finally, we predict the map segmentation from the fusion feature and add the supervision for this prediction to help the alignment module learning.

\section{Experiements}\label{sec:expriment}
\subsection{Implementation Details}
\subsubsection{Dataset}
We conduct extensive experiments on the OpenLane-V2\cite{wang2024openlane} dataset, which comprises 1000 scenarios, each approximately 15 seconds in duration. The training set covers around 27,000 frames, while the validation set contains approximately 4,800 frames. To augment the trajectory data, we extract trajectory data corresponding to the scenarios in OpenLaneV-2 based on the source id from the Argoverse 2 Motion Forecasting dataset. To ensure the accuracy of the trajectory data, we retain only the frames where the number of trajectories exceeded five times the number of centerlines. In 3D space, we retain all trajectories within the range of \([-50\, \text{m}, +50\, \text{m}]\) along the x-axis and \([-25\, \text{m}, +25\, \text{m}]\) along the y-axis as the raw data. 


\subsubsection{Metrics}

We adopt LaneSegNet \cite{li2023lanesegnet} as our baseline network and evaluate our model using metrics designed to capture lane detection accuracy and topological consistency. Specifically, we introduce the average precision \( AP_{\text{ls}} \), based on the distance \( D_{\text{ls}} \) between lane segments, and use \( TOP_{\text{lsls}} \) to assess the accuracy of topological connections between centerlines.

To further quantify prediction precision, we employ \( \text{AE}_{\text{type}} \) and \( \text{AE}_{\text{dist}} \) to measure attribute mismatches and spatial alignment with ground truth. The \(\text{AE}_{\text{type}} \) metric calculates the proportion of mismatched elements between two arrays. Meanwhile, \( \text{AE}_{\text{dist}} \) is computed by taking the mean minimum distance from predicted to ground truth points and vice versa, averaging these values for a comprehensive spatial accuracy measure. Higher values in either metric indicate a greater divergence from actual data. 
\subsubsection{Training Details}
We conduct our model training and baseline reproduction using 8 NVIDIA RTX A6000 GPUs. The model is trained for a total of 24 epochs and We employ the AdamW optimizer\cite{loshchilov2017decoupled} along with a cosine annealing schedule, with the initial learning rate set to 2e-4. The learning rate ratio is set to $10^{-3}$, and the batch size is 8.
Regarding the loss, the overall loss is defined as:

\begin{equation}
\begin{split}
\mathcal{L} = & \ \lambda_{\text{cls}} \mathcal{L}_{\text{cls}} + \lambda_{\text{vec}} \mathcal{L}_{\text{vec}} + \lambda_{\text{type}} \mathcal{L}_{\text{type}} \\
& + \lambda_{\text{insmask}} \mathcal{L}_{\text{insmask}}  + \lambda_{\text{bevseg}} \mathcal{L}_{\text{bevseg}}
+ \lambda_{\text{top}} \mathcal{L}_{\text{top}}
,
\end{split}
\end{equation}


where $\mathcal{L}_{\text{bevseg}}$ represents a novel BEV segmentation supervision term introduced in the alignment module, while the other defined losses are the same as those in LaneSegNet\cite{li2023lanesegnet}.
The hyperparameters are defined as follows:
$\lambda_{\text{cls}} = 1.5$, $\lambda_{\text{vec}} = 0.025$,
$\lambda_{\text{type}} = 0.1$, $\lambda_{\text{insmask}} = 3.0$,
$\lambda_{\text{bevseg}} = 3.0$,
$\lambda_{\text{top}} = 5.0$,
.

\subsection{Comparison with state-of-the-art}
To validate the effectiveness of the trajectory data, we reproduce the state-of-the-art SDMap prior fusion algorithm based on the LaneSegNet framework. Building on this foundation, we further introduced trajectory data to enhance model performance, thereby demonstrating the effectiveness of the trajectory data. 

As shown in Table \ref{traj fusion}, by incorporating the SDMap and trajectory data as prior information using our model, we achieve a 10.0 improvement in $AP_{\text{ls}}$ metric and  9.25 improvement in the $TOP_{\text{lsls}}$ metric compared with other prior model. Additionally, the $AE_{\text{type}}$ and $AE_{\text{dist}}$ metrics were significantly reduced.


Besides, Concurrent studies\cite{roddick2020predicting,yuan2024streammapnet,lilja2024localization} reveal a 54\% geographic overlap between Argoverse2's training and validation sets . Such spatial redundancy risks enabling model memorization of location-map pairs, artificially inflating validation metrics.
To rigorously assess generalization in unseen environments, we re-evaluated our method and baselines on OpenLane-V2 using StreamMapNet's\cite{yuan2024streammapnet} partitioning protocol. As shown in Table \ref{traj fusion}, our model achieves superior cross-scenario robustness.

\begin{table}[h]
    \centering
    \caption{Comparison with state-of-the-art, * means Validation on new split\cite{yuan2024streammapnet} of OpenLane-V2}
    \renewcommand{\arraystretch}{1.5}
    \setlength{\arrayrulewidth}{0.5mm}
    \label{traj fusion}
    \scalebox{0.8}{
    \begin{tabular}{cccccc}
    \hline
     Method & Extra info & $AP_{ls}$ & $AE_{type} \downarrow$ & $AE_{dist}\downarrow$ & $TOP_{lsls}$ \\ \hline
        LanesegNet\cite{li2023lanesegnet} & —— & 32.30 & 9.20 & 0.673 & 25.40 \\ 
         %
         MapTR\cite{liao2023maptrstructuredmodelinglearning} & —— & 27.00 & -- & 0.695 & -- \\ 
        MapTRv2\cite{liao2024maptrv2} & —— & 28.50 & -- & 0.702 & -- \\ 
        P-MapNet\cite{jiang2024p} & Ras SDMap & 34.66 &8.98 & 0.664 & 27.32 \\ 
        SMERF\cite{luo2024augmenting} & Vec SDMap  & 34.70 & 8.66 & 0.655 & 30.19 \\ 
        MapEx\cite{sun2023mind} & Vec SDMap  & 32.64 & 9.02 & 0.671 & 27.28 \\ 
        TrajTopo(ours) & SDMap/Trajectory & \textbf{42.30} & \textbf{7.58} & \textbf{0.618} & \textbf{34.65} \\
        \noalign{\global\arrayrulewidth=0.3mm}\hline
        \noalign{\global\arrayrulewidth=0.5mm}
        LaneSegNet*\cite{li2023lanesegnet} & —— & 16.67 & 18.72 & 0.7096 & 17.08  \\
        TrajTopo(ours)* & SDMap/Trajectory & \textbf{22.50} & \textbf{17.06} & \textbf{0.7033} & \textbf{21.22} \\ 
        \hline
    \end{tabular}
    }
\end{table}


\subsection{Ablation Study}
\subsubsection{Ablations on Additional module}
 To improve the overall metrics in our model, besides the modification mentioned in the model section, we also add SDMap prior using its optimal fusion strategy in other paper and align module.
 Moreover, to enhance topological reasoning capabilities, we integrate the topological reasoning module of LaneDAG into the baseline network. Table \ref{addition} shows the results of the experiment, we can see that the trajectory information as a standalone prior outperforms all methods utilizing SD Map prior across all evaluation metrics, illustrating the effectiveness of trajectory prior.

 \begin{table}[!ht]
    \centering
    \caption{Ablations on Additional module. * means baseline with LaneDAG Head\cite{jia2024lanedag}}
    \renewcommand{\arraystretch}{1.5}
    \setlength{\arrayrulewidth}{0.5mm}
    \label{addition}
    \scalebox{0.8}{
    \begin{tabular}{ccccc}
    \hline
    Method & $AP_{ls}$ & $AE_{type}\downarrow$ & $AE_{dist}\downarrow$ & $TOP_{lsls}$ \\ \hline
        Baseline & 32.30 & 9.20 & 0.673 & 25.40 \\ 
        Baseline* & 32.30 & 9.20 & 0.666 & 27.00(+1.60) \\ 
        +SDMap & 37.30(+5.0) & 8.77 & 0.638 & 31.10(+5.70) \\ 
        +Trajectory & 39.10(+6.8) & 7.87 & 0.621 & 31.18(+5.78) \\ 
        +Trajectory + SDMap & 42.30(+10.0) & 7.58 & 0.618 & 34.65(+9.25) \\ \hline
    \end{tabular}
    }
\end{table} 
\subsubsection{Ablations on Trajectory Fusion}
We propose two methods for encoding and fusing trajectory data, aiming to explore the optimal strategy for integrating trajectory data. To this end, we conduct a series of comparative experiments. As shown in Table \ref{ablation:traj}, different combinations of encoding and fusion methods all enhance the model's accuracy. In particular, the method of enhancing Bird's Eye View (BEV) feature using heatmaps achieves the best performance in topological reasoning tasks. Additionally, we find that vectorizing the trajectory data through KMeans clustering based on heatmaps and initializing the queries in the decoder also achieves similar performance improvements, although there is a slight decrease in topological reasoning. 

\begin{table*}[!ht]
    \centering
    \caption{Ablation study on trajectory fusion methods}
    \label{ablation:traj}
    \renewcommand{\arraystretch}{1.5}
    \setlength{\arrayrulewidth}{0.5mm}
    \begin{tabular*}{\textwidth}{@{\extracolsep{\fill}}%
        >{\centering\arraybackslash}p{2cm}
        >{\centering\arraybackslash}p{2cm}
        >{\centering\arraybackslash}p{2cm}
        >{\centering\arraybackslash}p{2cm}
        >{\centering\arraybackslash}p{2cm}
        >{\centering\arraybackslash}p{2cm}
        >{\centering\arraybackslash}p{2cm}
    }
    \hline
    \textbf{Heatmap} & \textbf{Far Sampling} & \textbf{KMeans Cluster} & $AP_{ls}$ & $AE_{type}$ $\downarrow$ & $AE_{dist}$$\downarrow$ & $TOP_{lsls}$\\ \hline
    Feat Aug & --- & --- & 42.30 & 7.58 & 0.618 & \textbf{34.65} \\ 
    --- & Query Init & --- & 39.65 & 7.99 & 0.637 & 31.31 \\ 
    --- & --- & Query Init & 38.56 & 7.75 & 0.635 & 31.30 \\ 
    Feat Aug & --- & Query Init & \textbf{42.41} & \textbf{7.43} & \textbf{0.613} & 33.81 \\ 
    Feat Aug & Query Init & --- & 40.68 & 7.46 & 0.614 & 33.02 \\ 
    \hline
    \end{tabular*}
\end{table*}

\subsubsection{Ablations on Fusion Alignment Module}

Table \ref{align module} presents the results of the ablation study on the fusion alignment module. The experimental results indicate that the fusion alignment module we proposed can effectively mitigate the spatial and accuracy biases introduced by prior information. 

\begin{table}[!h]
    \centering
    \caption{ Ablation study on Fusion Integrity Monitor}
    \renewcommand{\arraystretch}{1.5}
    \setlength{\arrayrulewidth}{0.5mm}
    \scalebox{0.9}{
    \begin{tabular}{ccccccc}
    \hline
        Method & $AP_{ls}$ & $AE_{type}\downarrow$ & $AE_{dist}\downarrow$ & $TOP_{lsls}$ \\ \hline
        with align module & \textbf{42.30} & \textbf{7.58} & \textbf{0.6175} & \textbf{34.65} \\ 
        w/o align module & 41.13 & 7.70 & 0.6204 & 34.18 \\ \hline
    \end{tabular}
    \label{align module}
    }
\end{table}

\begin{figure*}[b!]
    \centering
    \includegraphics[width=1\textwidth]{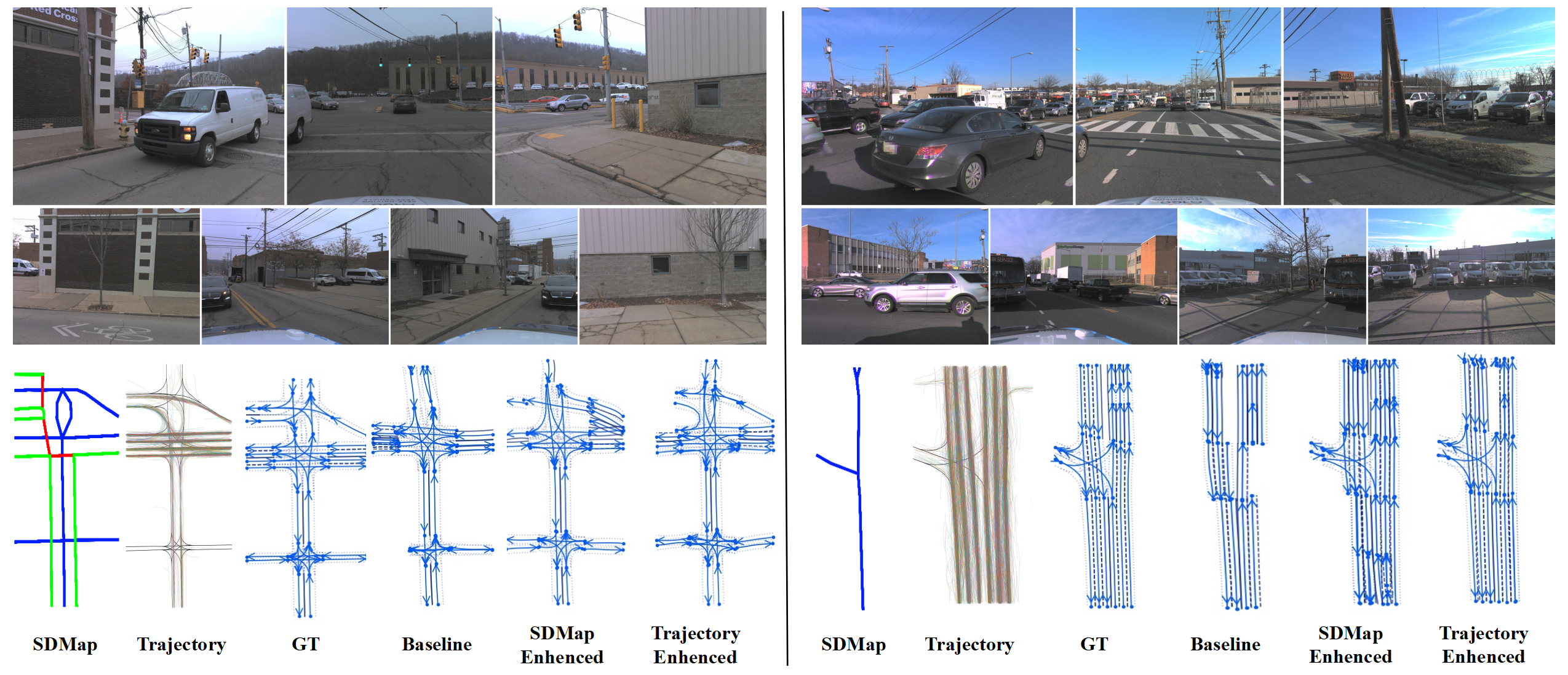}
    \caption{\textbf{Qualitative results}:It can be observed that under certain complex road conditions, the coarse sdmap provides some prior information, but its simple structure does not guarantee that the model has a more comprehensive understanding of the road scenario. In contrast, trajectory priors, due to their closer alignment with real-world scenarios, achieve better supplementary effects}
    \label{fig:visualize}
\end{figure*}
\subsubsection{Ablations on BEV Seg loss weight}
 For the new BEV Seg loss, we conducted several control experiments to assess different loss weights. The results are presented in Table\ref{bev seg loss}. According to the table, we finally choose to use 3.0 as the weight.
 \begin{table}[!h]
    \centering
    \caption{ Ablation study on BEV Seg loss weight}
    \renewcommand{\arraystretch}{1.5}
    \setlength{\arrayrulewidth}{0.5mm}
    \scalebox{0.9}{
    \begin{tabular}{ccccccc}
    \hline
        BEV Seg Loss Weight & $AP_{ls}$ & $AE_{type}\downarrow$ & $AE_{dist}\downarrow$ & $TOP_{lsls}$ \\ 
        \hline
        2.0 & \textbf{42.39} & \textbf{7.36} & 0.6196 & 34.36 \\ 
      
        3.0 & 42.30 & 7.58 & 0.6175 & \textbf{34.65} \\ 
        4.0 & 41.59 & 7.41 & \textbf{0.6160} & 34.47 \\ \hline
    \end{tabular}
    \label{bev seg loss}
    }
\end{table}
\subsection{Qualitative Visualization}
We conduct a visualization analysis of the prediction results, shown in Figure \ref{fig:visualize}. In the left scenario, trajectory enhancement yields clearer predictions compared to the baseline and SDMap-enhanced methods. At this complex intersection, vehicles cannot observe the merging lane on the front-right side, creating a blind spot. Both SDMap and trajectory data provide prior information to enhance the vehicle's situational awareness. However, predictions based on SDMap enhancement show larger deviations, with a tendency toward over-detection. In contrast, trajectory-enhanced predictions are more accurate and adhere more closely to the real road environment, reducing false detections and improving alignment with actual traffic flow. In the right scenario, SDMap provides only coarse prior information due to its simplistic structure, which limits its accuracy in complex road environments. In contrast, trajectory priors capture finer spatial and temporal details, enhancing model reliability and alignment with real traffic conditions.


\section{Conclusion}
In this paper, we introduce an innovative approach that leverages trajectory data as prior information to enhance lane segmentation perception and topology reasoning. To this end, we create a complementary dataset for the OpenLane-V2 dataset, specifically tailored to incorporate trajectory data. We then devise various fusion schemes and identify the optimal fusion strategy that best suits our purposes.

Extensive experiments have shown that the integration of trajectory data and our proposed method significantly improves the performance of existing models. We are confident that this novel use of prior information will provide a fresh perspective for future high-definition (HD) map construction tasks. 




\vfill

\bibliographystyle{ieeetr}
\bibliography{ref}

\end{document}